\documentclass[11pt]{article}

\usepackage{acl}

\usepackage{latexsym}
\usepackage[T1]{fontenc}
\usepackage{microtype}
\usepackage{graphicx}
\usepackage{booktabs}
\usepackage{tabularx}
\usepackage{multirow}
\usepackage{amsmath}
\usepackage[table]{xcolor}

\usepackage{fontspec}
\newfontfamily\jpfont{NotoSansCJKjp-Regular}[
    Path = fonts/,
    Extension = .otf,
    ItalicFont = NotoSansCJKjp-Regular,
    ItalicFeatures = {FakeSlant=0.2}
]
\newcommand{\jp}[1]{{\jpfont #1}}

\newcommand{\dratio}{\textit{DR}}

\title{Not All Irregularity Is Equal: Causally Isolating a Rare Failure Mode in Japanese Morphological Inflection}

\author{Wen Zhang \\
  \texttt{wenzhang0222@gmail.com}}

\begin{document}
\maketitle

\begin{abstract}
Neural morphological generation systems often achieve high aggregate accuracy on benchmark datasets, yet such performance can conceal systematic errors clustered in rare morphological subclasses. We present an orthography-aware diagnosis of Japanese past-tense verb inflection, treating \textit{hiragana} not merely as a transcriptional medium but as a representational system that encodes morphophonological structure. Using two character-level Transformer architectures evaluated across five random seeds, we show that although both systems exceed 97\% aggregate accuracy, a single structurally specific irregular subtype, verbs whose stems end in /e/ and require gemination before the past-tense suffix and make up fewer than 1\% of the data, accounts for a disproportionate 30--43\% share of residual errors and contributes roughly 34--48$\times$ its prevalence to total errors. We then move from diagnosis to causal isolation: controlled ablation experiments show that removing this subtype alone produces larger accuracy gains than removing all irregular verbs combined. These findings indicate that error concentration in neural morphological learning is not driven by irregularity \textit{per se}, but by the interaction between extreme low-frequency morphological patterns and specific orthographic processes. We argue that morphological evaluation should incorporate fine-grained subclass analysis, and discuss implications for data-efficient, developmentally plausible language model pretraining.%
\footnote{Code and data available at 
\url{https://github.com/wenzhang0222/not-all-irregularity}.}
\end{abstract}

\section{Introduction}

Japanese verbs are written in a hybrid system of \textit{kanji}, \textit{hiragana}, and \textit{katakana}. While verb stems may appear in either \textit{kanji} or \textit{hiragana}, inflectional suffixes are consistently marked in \textit{hiragana}. For example, the verb \textit{\jp{書く}}/\textit{\jp{かく}} \textit{kaku} becomes \textit{\jp{書いた}}/\textit{\jp{かいた}} \textit{kaita} in the past tense, where \textit{\jp{た}} \textit{ta} marks the past-tense suffix. \textit{Hiragana} is a moraic script, representing phonological timing units and marking processes such as gemination (the small \textit{\jp{っ}} \textit{tsu}) and vowel lengthening that are central to Japanese morphophonology \citep{kubozono2009, labrune2012}. Following the view that writing systems are structured representational systems rather than transparent encodings of speech \citep{sproat2000, danielsbright1996}, we treat \textit{hiragana} as a linguistically meaningful layer that may shape neural model generalization, and use it to eliminate confounds from \textit{kanji} homography by restricting all forms to \textit{hiragana}.

Most evaluations of Japanese morphological generation report aggregate exact-match accuracy \citep{cotterell-etal-2016-sigmorphon, vylomova-etal-2020-sigmorphon, goldman-etal-2023-sigmorphon}, often exceeding 95\% for high-resource languages such as Japanese. However, high aggregate accuracy does not necessarily indicate robust generalization, and can obscure systematic weaknesses concentrated in rare morphological subclasses \citep{goldman-etal-2022-un}. This is a manifestation of the ``long tail'' challenge in morphological learning \citep{cotterell-etal-2018-conll}, where models favor high-frequency regularities over infrequent but productive rules.

In this paper we combine two complementary methodologies to study this phenomenon in Japanese past-tense inflection. First, we conduct an orthography-aware \textbf{diagnosis}, using a fine-grained error taxonomy and subgroup-level accuracy analysis, aggregated across five random seeds and two architectures, to characterize \textit{where} models fail. Second, we conduct a \textbf{causal isolation} study, using controlled ablation experiments that selectively remove irregular verb subtypes, to test \textit{whether} the identified subtype is actually responsible for destabilizing generalization, and not irregularity in general. Combining them lets us both locate the failure and test its cause.

Our contributions are:
\begin{itemize}
\item A structural subgroup evaluation framework for Japanese morphological inflection, combining a six-category orthography-aware error taxonomy with a Disparity Ratio metric for quantifying subgroup-level error concentration.
\item A multi-seed, two-architecture empirical analysis showing that a single low-frequency irregular subtype (\textless 1\% of the data) accounts for 30--43\% of residual errors.
\item Controlled ablation experiments showing that selective removal of this subtype alone produces larger accuracy gains than removing all irregular verbs combined, suggesting that structural position, not frequency alone, is the primary driver of the effect.
\end{itemize}

\subsection{Japanese Past-Tense Morphology}

Japanese verbs are traditionally classified by inflectional behavior. Past-tense formation involves the suffix \textit{\jp{た}} \textit{-ta} and morphophonological alternations such as consonant mutation, gemination, and vowel changes, all of which are systematically reflected in \textit{hiragana} orthography. \textbf{\textit{Godan} (\textit{u}-) verbs} form the past tense through suffix-conditioned stem alternations, often involving consonant changes and gemination, e.g., \textit{\jp{かく}} \textit{kaku} `to write' $\to$ \textit{\jp{かいた}} \textit{kaita} `wrote'. \textbf{\textit{Ichidan} (\textit{ru}-) verbs} exhibit stable stems, with the past tense formed via direct suffix attachment, e.g., \textit{\jp{たべる}} \textit{taberu} `to eat' $\to$ \textit{\jp{たべた}} \textit{tabeta} `ate'. \textbf{Canonical irregular verbs} are a small closed class including \textit{\jp{する}} \textit{suru} `to do' $\to$ \textit{\jp{した}} \textit{shita} `did' and \textit{\jp{くる}} \textit{kuru} `to come' $\to$ \textit{\jp{きた}} \textit{kita} `came'. Because these alternations are encoded transparently in \textit{hiragana}, past-tense inflection is a dense testbed for studying morphophonological learning at the character level, without requiring lexical segmentation.

\section{Related Work}

\paragraph{Morphological Learning and Generalization.} Neural morphological inflection has been driven largely by Transformer-based encoder--decoder architectures \citep{vaswani2017}, achieving high aggregate accuracy on SIGMORPHON benchmarks \citep{vylomova-etal-2020-sigmorphon, goldman-etal-2023-sigmorphon}. There is growing recognition that aggregate metrics obscure performance disparities across morphological classes \citep{goldman-etal-2022-un}. We contribute to this line of work by showing how a single structurally idiosyncratic subclass can act as a stress test for neural generalization, and by testing this causally. This work synthesizes and extends our prior 
analyses of Japanese morphological inflection 
\citep{zhang-2026-moras, zhang-2026-irregularity}, combining orthography-aware error diagnosis with causal 
ablation methodology.

\paragraph{Subgroup Analysis and Fairness in NLP.} The challenge of modeling rare data is not unique to morphology; it mirrors broader concerns about subgroup disparity in neural models, where models often prioritize majority-group performance at the expense of underrepresented strata \citep{sagawa2020, pmlr-v81-buolamwini18a}. We adopt diagnostic tools from this literature, in particular a Disparity Ratio, to quantify error concentration across morphological subtypes, while noting that our ``subgroups'' are defined structurally and not socially \citep{blodgett-etal-2020-language}.

\paragraph{Inductive Bias and Irregularity.} The learnability of irregular morphology has long been central to debates between rule-based and connectionist accounts of language processing \citep{rumelhart-mcclelland-1986, pinker1994}. Our findings align with evidence that neural models can capture systematic patterns of morphological irregularity \citep{wu-etal-2019-morphological, ramarao-etal-2025-frequency}, but extend this literature by showing that not all irregular forms are equally destabilizing. Orthographic transparency and structural frequency, not irregular status alone, predict where models fail.

\paragraph{Evaluation Metrics and Orthographic Bias.}
Recent work shows that normalized crosslingual evaluation metrics introduce systematic biases 
rooted in tokenization, encoding, and orthographic differences \citep{yang-wilcox-arnett-2026}. Our findings offer a parallel concern at the subtype level within a single language: aggregate accuracy within Japanese morphological evaluation similarly obscures systematic failures clustered in structurally specific subclasses, suggesting that subgroup-aware diagnostics are necessary at both the crosslingual and intra-paradigm levels.

\section{Data}

We use a Japanese verb inflection dataset in standard morphological transduction format, following the setup of \citet{vylomova-etal-2020-sigmorphon} 
and \citet{goldman-etal-2023-sigmorphon}. All forms are converted to \textit{hiragana} to maintain orthographic consistency and eliminate confounds introduced by \textit{kanji} homography \citep{zhang-2023-pronunciation}. Each instance consists of three tab-separated fields: lemma, target form, and a placeholder tag (no explicit morphosyntactic features are provided, so the model must learn the lemma-to-form mapping directly), e.g.\ \textit{\jp{ねがえる}} \quad \textit{\jp{ねがえった}} \quad \_.

\subsection{Verb Classification}

Verbs are classified according to traditional Japanese conjugation classes, refined to capture orthography-sensitive variation. Canonical irregular verbs (\textit{\jp{する}}, \textit{\jp{くる}}; Type~3) and polysemous lemmas with multiple inflected forms are excluded to maintain a one-to-one lemma--form mapping. The remaining verbs fall into three types:

\begin{itemize}
\item \textbf{Type~1 (Godan)}: regular $u$-verbs, 
e.g.\ \textit{\jp{かく}} \textit{kaku} $\to$ 
\textit{\jp{かいた}} \textit{kaita}. Count: 2,503.
\item \textbf{Type~2 (Ichidan)}: regular $ru$-verbs, 
e.g.\ \textit{\jp{たべる}} \textit{taberu} $\to$ 
\textit{\jp{たべた}} \textit{tabeta}. Count: 1,298.
\item \textbf{Type~4} (irregular): further subdivided 
into three subtypes:
  \begin{itemize}
  \item \textbf{Type~4-1}: stem-final /i/ + 
gemination, e.g.\ \textit{\jp{まじる}} \textit{majiru} $\to$ 
\textit{\jp{まじった}} \textit{majitta}. Count: 119.
\item \textbf{Type~4-2}: stem-final /e/ + 
gemination, e.g.\ \textit{\jp{あきれかえる}} \textit{akirekaeru} $\to$ 
\textit{\jp{あきれかえった}} \textit{akirekaetta}. Count: 37.
\item \textbf{Type~4-3}: localized idiosyncratic 
deviations, e.g.\ \textit{\jp{いく}} \textit{iku} $\to$ 
\textit{\jp{いった}} \textit{itta}. Count: 1.
  \end{itemize}
\end{itemize}

Table~\ref{tab:dataset} summarizes dataset composition. Type~4-2 is the smallest structurally coherent subclass, comprising only 0.9\% of the data; the full verb list appears in Appendix~\ref{app:type42}.

\begin{table}[t]
\centering
\small
\begin{tabular}{lrr}
\toprule
\textbf{Verb Type} & \textbf{Count} & \textbf{Prop.\ (\%)} \\
\midrule
All verbs & 3{,}958 & 100 \\
Type 1 (Godan) & 2{,}503 & 63.2 \\
Type 2 (Ichidan) & 1{,}298 & 32.8 \\
Type 4 (irregular) & 157 & 4.0 \\
\quad 4-1 (/i/ + gemination) & 119 & 3.0 \\
\quad 4-2 (/e/ + gemination) & 37 & 0.9 \\
\quad 4-3 (localized) & 1 & 0.03 \\
\bottomrule
\end{tabular}
\caption{Dataset statistics by verb type. Canonical irregulars (\jp{する}, \jp{くる}; Type 3) are excluded, as described in \S3.1.}
\label{tab:dataset}
\end{table}

\section{Models}
We evaluate two character-level Transformer encoder--decoder architectures for Japanese past-tense inflection, both operating over \textit{hiragana} strings and generating inflected forms autoregressively. The first is a \textbf{Pointer-Generator Transformer} (henceforth \textbf{PGT}), incorporating a pointer-generator copying mechanism \citep{see-etal-2017-get} with attention over the input lemma, trained and evaluated on the morphological 
inflection data \citep{vylomova-etal-2020-sigmorphon}. The second follows the architecture of \citet{goldman-etal-2023-sigmorphon}, a \textbf{Lemma-Split Transformer} (henceforth \textbf{LST}) that additionally uses a lemma-split training and evaluation regime, preventing a lemma from appearing in both training and test data and improving generalization to unseen lemmas.

\section{Experimental Setup}

\subsection{Training Regime}
Both models are trained with cross-entropy loss and teacher forcing, using the default hyperparameter configurations from their respective shared-task baselines \citep{vylomova-etal-2020-sigmorphon, goldman-etal-2023-sigmorphon}, optimized with Adam \citep{kingma2015} and standard Transformer learning-rate scheduling \citep{vaswani2017}.

\subsection{Multi-Seed Baseline Evaluation}
Neural model performance is sensitive to random initialization, data shuffling, and optimization dynamics \citep{reimers-gurevych-2017-reporting}. To reduce 
selective-reporting bias, we train each model on the full dataset five times with different random seeds, using an 80/10/10 train/development/test split. The error taxonomy and quantitative 
error-distribution analyses (\S6.2--6.4) are aggregated across all five seeds. The controlled ablation experiments (\S6.7) involve eight training conditions per model and are conducted on the primary seed.

\subsection{Controlled Ablation Conditions}
To causally test the contribution of individual irregular subtypes, we train both architectures under eight controlled data conditions: the full dataset; regular verbs only (Types 1--2); the full dataset minus one irregular subtype (removing 4-1, 4-2, or 4-3 individually); and the full dataset minus each pairwise combination of subtypes. For every condition, the same verb types are removed from both training and test data.

\subsection{Evaluation Metrics}
We report \textbf{exact-match accuracy} at the lemma level \citep{cotterell-etal-2017-conll, goldman-etal-2023-sigmorphon}; \textbf{subgroup accuracy}, computed separately per verb type \citep{kann-schutze-2016-single, makarov-clematide-2018-imitation, vylomova-etal-2020-sigmorphon}; and the \textbf{Disparity Ratio} for a subgroup $g$,
\[
\dratio_g = \frac{\text{Error Share}_g}{\text{Data Share}_g},
\]
where a value greater than 1 indicates a disproportionately high error burden relative to prevalence, following diagnostic tools from subgroup fairness analysis \citep{sagawa2020, pmlr-v81-buolamwini18a, blodgett-etal-2020-language}.

\section{Results}

\subsection{Baseline Performance}
On the primary run, both systems achieve high aggregate accuracy on the full dataset: 97.98\% (PGT) and 97.73\% (LST). Averaged across all five seeds, accuracy is 97.17\% $\pm$ 0.65 (PGT) and 96.97\% $\pm$ 0.82 (LST) (mean $\pm$ sample standard deviation), confirming that high aggregate accuracy is a stable property of both architectures. Despite this overall stability, errors cluster in specific low-frequency subclasses, as we show next.

\subsection{Error Taxonomy}
We defined a taxonomy of six orthography- and morphophonology-sensitive failure modes (Table~\ref{tab:taxonomy}): gemination omission, gemination insertion, phonological substitution, morpheme boundary errors, character recognition (UNK) errors, and compound verb structural errors. We used this taxonomy to manually classify all residual errors (Table~\ref{tab:quant}). This taxonomy highlights that residual failures are systematic, not random noise. For example, failures to insert, or spuriously insert, the small \textit{\jp{っ}} reflect specific difficulty with consonant doubling.

\begin{table*}[t]
\centering
\small
\begin{tabularx}{\textwidth}{l X l}
\toprule
\textbf{Error Type} & \textbf{Description} & \textbf{Orthographic/Phonological Property} \\
\midrule
Gemination omission & Failure to insert small \jp{っ} & Consonant doubling \\
Gemination insertion & Spurious insertion of \jp{っ} & Consonant doubling \\
Phonological substitution & Incorrect consonant or vowel in stem & Sound alternation \\
Morpheme boundary error & Misaligned suffix attachment & Boundary detection \\
Character recognition error & Generation of UNK symbol & Encoding/representation \\
Compound verb error & Structural collapse in compound verbs & Compound segmentation \\
\bottomrule
\end{tabularx}
\caption{Error taxonomy for \textit{hiragana}-based inflection.}
\label{tab:taxonomy}
\end{table*}

\subsection{Quantitative Error Distribution}
Aggregated across five seeds, we observe 104 total errors, 53 from the PGT system and 51 from the LST system. Table~\ref{tab:quant} reports the six observed error types. Gemination-related errors (omission and insertion combined) dominate residual failures, accounting for 75.5\% (PGT) and 80.3\% (LST) of all errors, with omissions far more frequent than insertions in both systems.

\begin{table}[t]
\centering
\small
\begin{tabularx}{\columnwidth}{>{\raggedright\arraybackslash}X>{\centering\arraybackslash}X>{\centering\arraybackslash}X}
\toprule
\textbf{Error Type} & \textbf{PGT} & \textbf{LST} \\
\midrule
Gemination omission & 33 (62.3\%) & 32 (62.7\%) \\
Gemination insertion & 7 (13.2\%) & 9 (17.6\%) \\
Phonological substitution & 6 (11.3\%) & 4 (7.8\%) \\
Morpheme boundary & 3 (5.7\%) & 4 (7.8\%) \\
Character recognition (UNK) & 1 (1.9\%) & 1 (2.0\%) \\
Compound verb error & 3 (5.7\%) & 1 (2.0\%) \\
\midrule
\textbf{Total} & \textbf{53} & \textbf{51} \\
\bottomrule
\end{tabularx}
\caption{Quantitative error distribution by error type.}
\label{tab:quant}
\end{table}

\begin{figure}[t]
\centering
\includegraphics[width=\columnwidth]{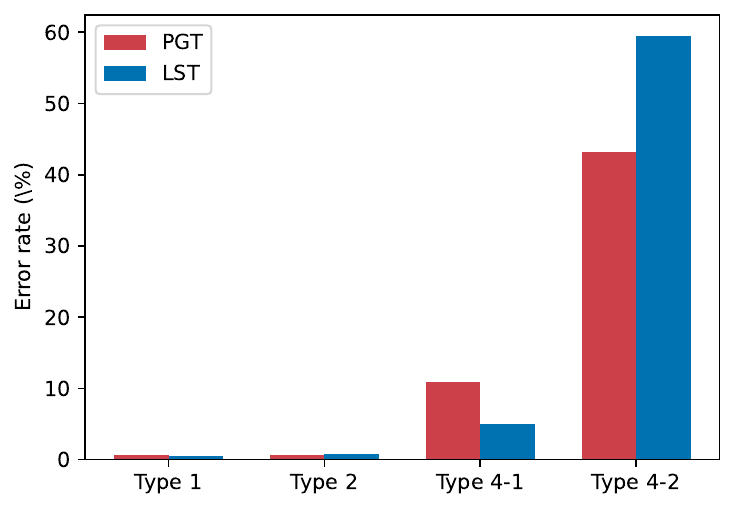}
\caption{Per-type error rate (errors / total items of that type) for PGT and LST systems. Type~4-2 shows an error rate of 43\% (PGT) and 60\% (LST), more than 70 times the rate of Type~1 verbs.}
\label{fig:errorrate}
\end{figure}

\subsection{Verb-Class Asymmetry and Disparity Ratios}
Table~\ref{tab:verbclass} breaks the same errors down by verb type. Type 4-2 verbs are massively overrepresented, accounting for 30.2\% (PGT) and 43.1\% (LST) of all errors while comprising just 0.9\% of the dataset. Type 4-1 is also somewhat overrepresented, while Type 1 and Type 2 show error rates below their dataset share. Figure~\ref{fig:errorrate} shows the per-type error rate directly, making the disproportionate difficulty of Type~4-2 visually apparent.

\begin{table}[t]
\centering
\small
\begin{tabularx}{\columnwidth}{>{\centering\arraybackslash}X>{\centering\arraybackslash}X>{\centering\arraybackslash}X>{\centering\arraybackslash}X}
\toprule
\textbf{Verb Type} & \textbf{PGT Err.} & \textbf{LST Err.} & \textbf{Data \%} \\
\midrule
Type 1 & 15 (28.3\%) & 13 (25.5\%) & 63.2 \\
Type 2 & 9 (17.0\%) & 10 (19.6\%) & 32.8 \\
Type 4-1 & 13 (24.5\%) & 6 (11.8\%) & 3.0 \\
Type 4-2 & \cellcolor{gray!25}\textbf{16 (30.2\%)} & \cellcolor{gray!25}\textbf{22 (43.1\%)} & \cellcolor{gray!25}\textbf{0.9} \\
Type 4-3 & 0 & 0 & 0.03 \\
\midrule
\textbf{Total} & \textbf{53} & \textbf{51} & \textbf{100} \\
\bottomrule
\end{tabularx}
\caption{Error distribution by verb type against 
dataset share.}
\label{tab:verbclass}
\end{table}

\begin{figure}[t]
\centering
\includegraphics[width=\columnwidth]{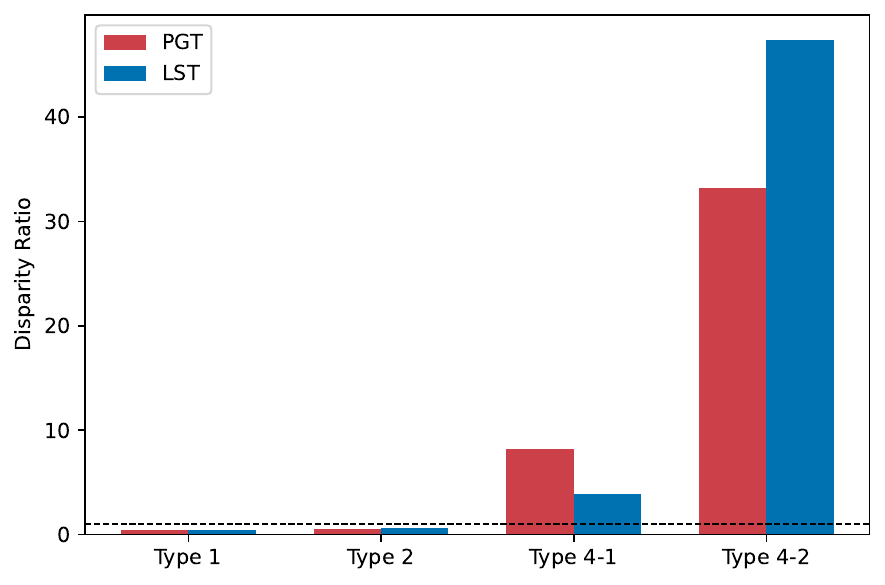}
\caption{Disparity Ratio by verb type for PGT and LST systems. Type~4-2 shows a dramatically higher disparity ratio than all other subtypes, contributing 34--48$\times$ its proportional 
share to total errors.}
\label{fig:disparity}
\end{figure}

Table~\ref{tab:disparity} converts this into Disparity Ratios. 
Type~4-2 contributes 34--48$\times$ its proportional representation to total errors, as visualized in Figure~\ref{fig:disparity}, the largest disparity of any subclass in either system. This pattern 
is consistent across both architectures.

To confirm this over-representation is not an artifact of small counts, we tested it directly. Under the null hypothesis that a verb's probability of erring is independent of its type (i.e., errors are distributed in proportion to each type's data share), we would expect Type 4-2 to account for only 0.9\% of errors. A one-sided exact binomial test rejects this null decisively for both systems (PGT: 16/53 errors, $p = 3.6\times10^{-20}$; LST: 22/51 errors, $p = 2.7\times10^{-31}$).

\begin{table}[t]
\centering
\small
\begin{tabular}{lrr}
\toprule
\textbf{Verb Type} & \textbf{\dratio\ (PGT)} & 
\textbf{\dratio\ (LST)} \\
\midrule
Type 1 & 0.45 & 0.40 \\
Type 2 & 0.52 & 0.60 \\
Type 4-1 & 8.17 & 3.93 \\
Type 4-2 & \textbf{33.6} & \textbf{47.9} \\
\bottomrule
\end{tabular}
\caption{Disparity Ratio (error share / data share) 
by verb type. Values $>$1 indicate error 
over-representation relative to prevalence.}
\label{tab:disparity}
\end{table}

\subsection{Cross-Model Consistency}
Despite architectural differences, error patterns are broadly consistent across the PGT and LST systems. Gemination-related failures dominate in both (75.5\%/80.3\%), Type 4-2 is the most overrepresented subclass in both (Table 4), and an identical single UNK error appears in both systems' outputs (\textit{\jp{つっぷした}} $\to$ \textit{\jp{っ}}\texttt{<UNK>}\textit{\jp{した}}). This consistency, together with the seed-level stability reported in \S6.1, suggests the failures reflect systematic properties of orthographic representation rather than idiosyncrasies of a single model.

\subsection{Qualitative Error Patterns}
The dominant Type 4-2 failure is omission of required gemination before a stem-final /e/, e.g.\ \textit{\jp{あきれかえった}} \textit{akirekaetta} $\to$ \textit{\jp{あきれかえた}} \textit{akirekaeta}. Spurious gemination insertion occurs mainly in Type 2 verbs, e.g.\ \textit{\jp{おきた}} \textit{okita} $\to$ \textit{\jp{おきった}} \textit{okitta}. Morpheme boundary errors arise when compound boundaries are unmarked in \textit{hiragana}, e.g.\ \textit{\jp{ほめたたえた}} \textit{hometataeta} $\to$ \textit{\jp{ほめたえた}} \textit{hometaeta}.

\subsection{Causal Ablation: Isolating Type 4-2}
\begin{figure}[t]
\centering
\includegraphics[width=\columnwidth]{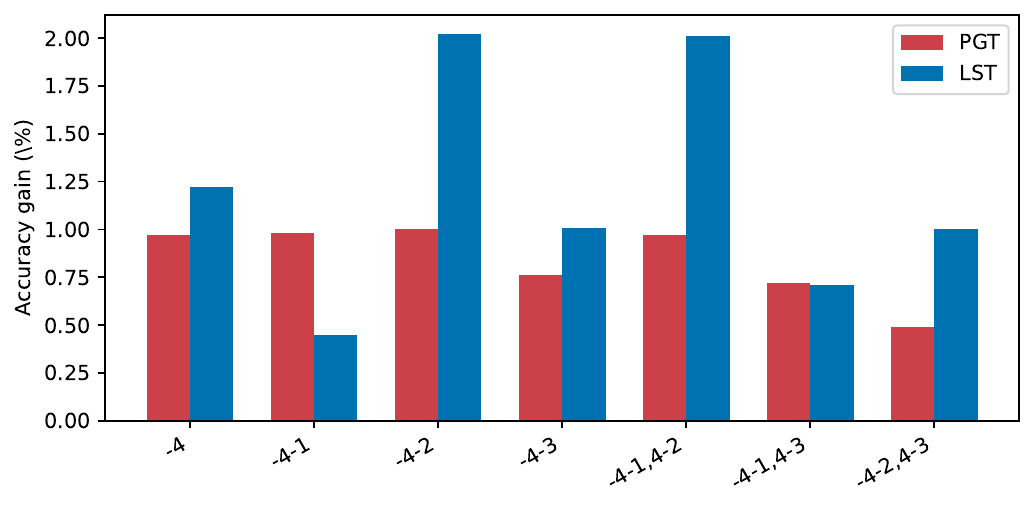}
\caption{Accuracy gains from Table~\ref{tab:gains} 
visualized. The $-$4-2 condition stands out clearly 
in both systems.}

\label{fig:ablation}
\end{figure}

The analyses above are descriptive: they show 
\textit{where} errors concentrate, but not whether 
removing the implicated subtype actually improves 
generalization more than removing irregularity 
broadly. Table~\ref{tab:gains} and 
Figure~\ref{fig:ablation} summarize the accuracy 
gain over the full-data baseline for each condition. 
Table~\ref{tab:ablation} reports test accuracy 
under all eight ablation conditions. Because the 
same verb types are removed from both training 
and test across all conditions, accuracy differences 
between conditions reflect differences in which 
subtypes destabilize learning of the remaining 
vocabulary, not differences in test set composition.

\begin{table}[t]
\centering
\small
\begin{tabular}{lrr}
\toprule
\textbf{Condition} & \textbf{PGT gain} & 
\textbf{LST gain} \\
\midrule
$-$4 (all irregulars) & +0.97 & +1.22 \\
$-$4-1 & +0.98 & +0.45 \\
$-$4-2 & \textbf{+1.00} & \textbf{+2.02} \\
$-$4-3 & +0.76 & +1.01 \\
$-$4-1,4-2 & +0.97 & +2.01 \\
$-$4-1,4-3 & +0.72 & +0.71 \\
$-$4-2,4-3 & +0.49 & +1.00 \\
\bottomrule
\end{tabular}
\caption{Accuracy gain over full-data baseline for 
each ablation condition. Boldface marks the 
single-subtype condition producing the largest 
improvement in each system.}
\label{tab:gains}
\end{table}

\begin{table}[t]
\centering
\small
\begin{tabular}{lrr}
\toprule
\textbf{Condition} & \textbf{PGT} & \textbf{LST} \\
\midrule
Full (1--4) & 97.98 & 97.73 \\
$-$4 (all irregulars) & 98.95 & 98.95 \\
$-$4-1 & 98.96 & 98.18 \\
$-$4-2 & \textbf{98.98} & \textbf{99.75} \\
$-$4-3 & 98.74 & 98.74 \\
$-$4-1,4-2 & 98.95 & 99.74 \\
$-$4-1,4-3 & 98.70 & 98.44 \\
$-$4-2,4-3 & 98.47 & 98.73 \\
\bottomrule
\end{tabular}
\caption{Test-set exact-match accuracy (\%) under 
each ablation condition. Bold marks the best 
condition per system.}
\label{tab:ablation}
\end{table}

For the \textbf{LST system}, removing only Type 4-2 yields the largest
observed gain: 97.73\% $\to$ 99.75\% (+2.02),
corresponding to an approximately 89\% reduction in error rate.
The gain is substantially larger than removing all irregular verbs
(+1.22) and matches or exceeds every other
single- or paired-subtype removal. For the \textbf{PGT system},
removing Type 4-2 alone is still the single best condition
(97.98\% $\to$ 98.98\%, +1.00).

In both systems, removing the full irregular set (Type 4) never yields the maximal accuracy. Some irregular subtypes (4-1, 4-3) can be retained without cost: in the LST system, removing only Type~4-2 while retaining 4-1 and 4-3 (condition $-$4-2) achieves 99.75\%, and additionally removing 4-1 as well (condition $-$4-1,4-2) performs nearly as well at 99.74\%. This indicates that the relevant factor is not irregular status \textit{per se}, but the specific structural configuration instantiated by Type 4-2.

\section{Discussion}

Taken together, the diagnostic and causal results support a consistent picture. Irregularity is not uniformly detrimental to neural morphological learning. A single low-frequency subtype, verbs requiring gemination after a stem-final /e/, is structurally distinct from other irregulars. It accounts for a disproportionate share of errors (34--48$\times$ its data share) and, for LST, is causally responsible for most of the achievable accuracy gain from irregular-verb ablation.

The picture is more nuanced for the PGT architecture. Type 4-2 removal is directionally consistent with the LST result, remaining the best single condition, but the margin over alternative ablations is small. We view this as a genuine and informative asymmetry and not a weakness to be smoothed over: the diagnostic finding, that Type 4-2 is where errors concentrate, is robust across architectures and seeds, while the strength of the causal evidence for removing Type 4-2 specifically varies between them.

Several factors likely contribute to Type 4-2's structural distinctiveness. First, consonant gemination (small \jp{っ}) introduces an additional mora explicitly represented in \textit{hiragana}, altering a word's rhythmic structure in a way character-level models must generate in the correct context \citep{pimentel-etal-2020-information}. Second, orthographic form alone does not encode all conditioning factors (e.g., lexical frequency), so some verbs remain more error-prone than surface similarity would suggest. Third, at under 1\% of training data, Type 4-2 instances are too sparse for models to reliably learn the conditioning environment, while still being frequent enough, and structurally regular enough, to be systematically attempted rather than simply memorized or ignored. This contrasts with Type 4-3, whose single instance produces zero errors in both systems, likely because it is either trivially memorized or entirely absent from test predictions in a way that does not surface as a systematic failure.

This has a direct implication for evaluation practice. Aggregate accuracy near 98\% in both our systems obscures an effect visible only under subtype-level analysis; we suggest the same is plausibly true of aggregate scores on existing benchmarks. Per-item predictions can be regrouped by structural subclass and scored with the Disparity Ratio metric introduced in \S5.4. Evaluation protocols for data-efficient pretraining may therefore benefit from subgroup-aware diagnostics of the kind proposed here, and not aggregate accuracy alone.

\paragraph{Future Directions.}
The structural confusability effect identified here raises two natural extensions. First, whether the same effect appears when models are evaluated zero-shot at scale, testing whether frontier language models show the same Type~4-2 sensitivity or whether scale eliminates the inductive bias. Second, whether the Disparity 
Ratio metric generalizes beyond Japanese to other morphologically rich languages, testing whether structural confusability constitutes a 
domain-general constraint on morphological learning.

\section{Conclusion}

We presented a combined diagnostic and causal analysis of Japanese past-tense inflection. Aggregating over five seeds and two architectures, we showed that a single irregular subtype comprising under 1\% of the data accounts for 30--43\% of residual errors (a 34--48$\times$ disparity ratio), concentrated in gemination-related failures. Controlled ablation experiments show that removing only this subtype improves generalization more than removing all irregular verbs. These results demonstrate that high aggregate accuracy can mask structurally localized failures, that not all irregularity is equally destabilizing to neural generalization, and that fine-grained, subgroup-aware evaluation is necessary to reveal and causally test such effects. We argue this evaluation practice is particularly relevant to sample-efficient, developmentally plausible language modeling. Future work will test whether the same structural confusability effects appear in large language models, asking whether scale alone is sufficient to overcome the inductive biases identified here.

\section*{Limitations}

Several limitations qualify our findings. First, our ablation experiments were each run once per condition; replication across multiple training runs would strengthen the causal claim. Second, our study focuses on a single language, a single morphological paradigm, and two Transformer-based architectures. Cross-linguistic validation is necessary to determine whether comparable low-frequency structural subtypes produce similar disparities in other morphologically rich languages, and whether larger pretrained models exhibit different sensitivity to the patterns we identify. Third, our analysis is correlational with respect to the specific mechanism (frequency vs.\ structural complexity vs.\ their interaction) underlying Type 4-2's distinctiveness; the ablation results support a causal role for this subtype's presence in training data, but do not isolate frequency from structural complexity as independent factors. We leave controlled frequency-matched comparisons, and evaluation under alternative input representations (e.g., IPA or romanization), to future work.

\section*{Ethics Statement}

This work involves no sensitive data, human subjects, or personally identifiable information. Data and models are derived from publicly available databases and SIGMORPHON shared-task resources.

\section*{Acknowledgments}
We thank the BabyLM 2026 workshop for their thoughtful feedback, which strengthened this work.

\clearpage
\bibliography{references}

@inproceedings{vylomova-etal-2020-sigmorphon,
    title = "{SIGMORPHON} 2020 Shared Task 0: Typologically Diverse Morphological Inflection",
    author = "Vylomova, Ekaterina  and
      White, Jennifer  and
      Salesky, Elizabeth  and
      Mielke, Sabrina J.  and
      Wu, Shijie  and
      Ponti, Edoardo Maria  and
      Maudslay, Rowan Hall  and
      Zmigrod, Ran  and
      Valvoda, Josef  and
      Toldova, Svetlana  and
      Tyers, Francis  and
      Klyachko, Elena  and
      Yegorov, Ilya  and
      Krizhanovsky, Natalia  and
      Czarnowska, Paula  and
      Nikkarinen, Irene  and
      Krizhanovsky, Andrew  and
      Pimentel, Tiago  and
      Torroba Hennigen, Lucas  and
      Kirov, Christo  and
      Nicolai, Garrett  and
      Williams, Adina  and
      Anastasopoulos, Antonios  and
      Cruz, Hilaria  and
      Chodroff, Eleanor  and
      Cotterell, Ryan  and
      Silfverberg, Miikka  and
      Hulden, Mans",
    editor = "Nicolai, Garrett  and
      Gorman, Kyle  and
      Cotterell, Ryan",
    booktitle = "Proceedings of the 17th SIGMORPHON Workshop on Computational Research in Phonetics, Phonology, and Morphology",
    month = jul,
    year = "2020",
    address = "Online",
    publisher = "Association for Computational Linguistics",
    url = "https://aclanthology.org/2020.sigmorphon-1.1/",
    doi = "10.18653/v1/2020.sigmorphon-1.1",
    pages = "1--39"
}

@inproceedings{goldman-etal-2023-sigmorphon,
    title = "{SIGMORPHON}{--}{U}ni{M}orph 2023 Shared Task 0: Typologically Diverse Morphological Inflection",
    author = "Goldman, Omer  and
      Batsuren, Khuyagbaatar  and
      Khalifa, Salam  and
      Arora, Aryaman  and
      Nicolai, Garrett  and
      Tsarfaty, Reut  and
      Vylomova, Ekaterina",
    editor = {Nicolai, Garrett  and
      Chodroff, Eleanor  and
      Mailhot, Frederic  and
      {\c{C}}{\"o}ltekin, {\c{C}}a{\u{g}}r{\i}},
    booktitle = "Proceedings of the 20th SIGMORPHON workshop on Computational Research in Phonetics, Phonology, and Morphology",
    month = jul,
    year = "2023",
    address = "Toronto, Canada",
    publisher = "Association for Computational Linguistics",
    url = "https://aclanthology.org/2023.sigmorphon-1.13/",
    doi = "10.18653/v1/2023.sigmorphon-1.13",
    pages = "117--125"
}

@inproceedings{cotterell-etal-2016-sigmorphon,
    title = "The {SIGMORPHON} 2016 Shared {T}ask{---}{M}orphological Reinflection",
    author = "Cotterell, Ryan  and
      Kirov, Christo  and
      Sylak-Glassman, John  and
      Yarowsky, David  and
      Eisner, Jason  and
      Hulden, Mans",
    editor = "Elsner, Micha  and
      Kuebler, Sandra",
    booktitle = "Proceedings of the 14th {SIGMORPHON} Workshop on Computational Research in Phonetics, Phonology, and Morphology",
    month = aug,
    year = "2016",
    address = "Berlin, Germany",
    publisher = "Association for Computational Linguistics",
    url = "https://aclanthology.org/W16-2002/",
    doi = "10.18653/v1/W16-2002",
    pages = "10--22"
}

@inproceedings{cotterell-etal-2017-conll,
    title = "{C}o{NLL}-{SIGMORPHON} 2017 Shared Task: Universal Morphological Reinflection in 52 Languages",
    author = {Cotterell, Ryan  and
      Kirov, Christo  and
      Sylak-Glassman, John  and
      Walther, G{\'e}raldine  and
      Vylomova, Ekaterina  and
      Xia, Patrick  and
      Faruqui, Manaal  and
      K{\"u}bler, Sandra  and
      Yarowsky, David  and
      Eisner, Jason  and
      Hulden, Mans},
    editor = "Hulden, Mans",
    booktitle = "Proceedings of the {C}o{NLL} {SIGMORPHON} 2017 Shared Task: Universal Morphological Reinflection",
    month = aug,
    year = "2017",
    address = "Vancouver",
    publisher = "Association for Computational Linguistics",
    url = "https://aclanthology.org/K17-2001/",
    doi = "10.18653/v1/K17-2001",
    pages = "1--30"
}

@inproceedings{cotterell-etal-2018-conll,
    title = "The {C}o{NLL}{--}{SIGMORPHON} 2018 Shared Task: Universal Morphological Reinflection",
    author = "Cotterell, Ryan  and
      Kirov, Christo  and
      Sylak-Glassman, John  and
      Walther, G{\'e}raldine  and
      Vylomova, Ekaterina  and
      McCarthy, Arya D.  and
      Kann, Katharina  and
      Mielke, Sabrina J.  and
      Nicolai, Garrett  and
      Silfverberg, Miikka  and
      Yarowsky, David  and
      Eisner, Jason  and
      Hulden, Mans",
    editor = "Hulden, Mans  and
      Cotterell, Ryan",
    booktitle = "Proceedings of the {C}o{NLL}{--}{SIGMORPHON} 2018 Shared Task: Universal Morphological Reinflection",
    month = oct,
    year = "2018",
    address = "Brussels",
    publisher = "Association for Computational Linguistics",
    url = "https://aclanthology.org/K18-3001/",
    doi = "10.18653/v1/K18-3001",
    pages = "1--27"
}

@inproceedings{kann-schutze-2016-single,
    title = "Single-Model Encoder-Decoder with Explicit Morphological Representation for Reinflection",
    author = {Kann, Katharina  and
      Sch{\"u}tze, Hinrich},
    editor = "Erk, Katrin  and
      Smith, Noah A.",
    booktitle = "Proceedings of the 54th Annual Meeting of the Association for Computational Linguistics (Volume 2: Short Papers)",
    month = aug,
    year = "2016",
    address = "Berlin, Germany",
    publisher = "Association for Computational Linguistics",
    url = "https://aclanthology.org/P16-2090/",
    doi = "10.18653/v1/P16-2090",
    pages = "555--560"
}

@inproceedings{wu-etal-2019-morphological,
    title = "Morphological Irregularity Correlates with Frequency",
    author = "Wu, Shijie  and
      Cotterell, Ryan  and
      O{'}Donnell, Timothy",
    editor = "Korhonen, Anna  and
      Traum, David  and
      M{\`a}rquez, Llu{\'i}s",
    booktitle = "Proceedings of the 57th Annual Meeting of the Association for Computational Linguistics",
    month = jul,
    year = "2019",
    address = "Florence, Italy",
    publisher = "Association for Computational Linguistics",
    url = "https://aclanthology.org/P19-1505/",
    doi = "10.18653/v1/P19-1505",
    pages = "5117--5126"
}

@inproceedings{makarov-clematide-2018-imitation,
    title = "Imitation Learning for Neural Morphological String Transduction",
    author = "Makarov, Peter  and
      Clematide, Simon",
    editor = "Riloff, Ellen  and
      Chiang, David  and
      Hockenmaier, Julia  and
      Tsujii, Jun{'}ichi",
    booktitle = "Proceedings of the 2018 Conference on Empirical Methods in Natural Language Processing",
    month = oct # "-" # nov,
    year = "2018",
    address = "Brussels, Belgium",
    publisher = "Association for Computational Linguistics",
    url = "https://aclanthology.org/D18-1314/",
    doi = "10.18653/v1/D18-1314",
    pages = "2877--2882"
}

@inproceedings{ramarao-etal-2025-frequency,
    title = "Frequency matters: Modeling irregular morphological patterns in {S}panish with Transformers",
    author = "Kakolu Ramarao, Akhilesh  and
      Tang, Kevin  and
      Baer-Henney, Dinah",
    editor = "Che, Wanxiang  and
      Nabende, Joyce  and
      Shutova, Ekaterina  and
      Pilehvar, Mohammad Taher",
    booktitle = "Findings of the Association for Computational Linguistics: ACL 2025",
    month = jul,
    year = "2025",
    address = "Vienna, Austria",
    publisher = "Association for Computational Linguistics",
    url = "https://aclanthology.org/2025.findings-acl.230/",
    doi = "10.18653/v1/2025.findings-acl.230",
    pages = "4474--4489",
    ISBN = "979-8-89176-256-5"
}

@inproceedings{vaswani2017,
  title     = {Attention Is All You Need},
  author    = {Vaswani, Ashish and Shazeer, Noam and Parmar, Niki and Uszkoreit, Jakob and Jones, Llion and Gomez, Aidan N. and Kaiser, {\L}ukasz and Polosukhin, Illia},
  year      = {2017},
  booktitle = {Advances in Neural Information Processing Systems},
  volume    = {30},
  pages     = {5998--6008}
}

@inproceedings{kingma2015,
  title     = {Adam: A Method for Stochastic Optimization},
  author    = {Kingma, Diederik P. and Ba, Jimmy},
  year      = {2015},
  booktitle = {International Conference on Learning Representations (ICLR)}
}

@inproceedings{reimers-gurevych-2017-reporting,
    title = "Reporting Score Distributions Makes a Difference: Performance Study of {LSTM}-networks for Sequence Tagging",
    author = "Reimers, Nils  and
      Gurevych, Iryna",
    editor = "Palmer, Martha  and
      Hwa, Rebecca  and
      Riedel, Sebastian",
    booktitle = "Proceedings of the 2017 Conference on Empirical Methods in Natural Language Processing",
    month = sep,
    year = "2017",
    address = "Copenhagen, Denmark",
    publisher = "Association for Computational Linguistics",
    url = "https://aclanthology.org/D17-1035/",
    doi = "10.18653/v1/D17-1035",
    pages = "338--348"
}

@inproceedings{sagawa2020,
  title     = {Distributionally Robust Neural Networks for Group Shifts: On the Importance of Regularization for Worst-Case Generalization},
  author    = {Sagawa, Shiori and Koh, Pang Wei and Hashimoto, Tatsunori B. and Liang, Percy},
  year      = {2020},
  booktitle = {International Conference on Learning Representations (ICLR)}
}

@InProceedings{pmlr-v81-buolamwini18a,
  title = 	 {Gender Shades: Intersectional Accuracy Disparities in Commercial Gender Classification},
  author = 	 {Buolamwini, Joy and Gebru, Timnit},
  booktitle = 	 {Proceedings of the 1st Conference on Fairness, Accountability and Transparency},
  pages = 	 {77--91},
  year = 	 {2018},
  editor = 	 {Friedler, Sorelle A. and Wilson, Christo},
  volume = 	 {81},
  series = 	 {Proceedings of Machine Learning Research},
  month = 	 {23--24 Feb},
  publisher =    {PMLR},
  url = 	 {https://proceedings.mlr.press/v81/buolamwini18a.html}
}

@inproceedings{blodgett-etal-2020-language,
    title = "Language (Technology) is Power: A Critical Survey of ``Bias'' in {NLP}",
    author = "Blodgett, Su Lin  and
      Barocas, Solon  and
      Daum{\'e} III, Hal  and
      Wallach, Hanna",
    editor = "Jurafsky, Dan  and
      Chai, Joyce  and
      Schluter, Natalie  and
      Tetreault, Joel",
    booktitle = "Proceedings of the 58th Annual Meeting of the Association for Computational Linguistics",
    month = jul,
    year = "2020",
    address = "Online",
    publisher = "Association for Computational Linguistics",
    url = "https://aclanthology.org/2020.acl-main.485/",
    doi = "10.18653/v1/2020.acl-main.485",
    pages = "5454--5476"
}

@incollection{rumelhart-mcclelland-1986,
  author    = {Rumelhart, David E. and McClelland, James L.},
  title     = {On Learning the Past Tenses of English Verbs},
  booktitle = {Parallel Distributed Processing: Explorations in the Microstructure of Cognition},
  editor    = {McClelland, James L. and Rumelhart, David E. and the PDP Research Group},
  volume    = {2},
  pages     = {216--271},
  publisher = {MIT Press},
  year      = {1986}
}

@book{pinker1994,
  title     = {The Language Instinct},
  author    = {Pinker, Steven},
  year      = {1994},
  publisher = {William Morrow and Company}
}

@incollection{kubozono2009,
    author = "Kubozono, Haruo and Ito, Junko and Mester, Armin",
    title = "Consonant Gemination in Japanese Loanword Phonology",
    booktitle = "Current Issues in Unity and Diversity of Languages:
                 Collection of Papers Selected from the 18th International
                 Congress of Linguists [CIL 18]",
    editor = "The Linguistic Society of Korea",
    pages = "953--973",
    publisher = "Dongam Publishing Co.",
    address = "Republic of Korea",
    year = "2009"
}

@book{labrune2012,
  title     = {The Phonology of {J}apanese},
  author    = {Labrune, Laurence},
  year      = {2012},
  publisher = {Oxford University Press}
}

@book{sproat2000,
  title     = {A Computational Theory of Writing Systems},
  author    = {Sproat, Richard},
  year      = {2000},
  publisher = {Cambridge University Press}
}

@book{danielsbright1996,
  title     = {The World's Writing Systems},
  editor    = {Daniels, Peter T. and Bright, William},
  year      = {1996},
  publisher = {Oxford University Press},
  address   = {Oxford}
}

@inproceedings{pimentel-etal-2020-information,
    title = "Information-Theoretic Probing for Linguistic Structure",
    author = "Pimentel, Tiago  and
      Valvoda, Josef  and
      Maudslay, Rowan Hall  and
      Zmigrod, Ran  and
      Williams, Adina  and
      Cotterell, Ryan",
    editor = "Jurafsky, Dan  and
      Chai, Joyce  and
      Schluter, Natalie  and
      Tetreault, Joel",
    booktitle = "Proceedings of the 58th Annual Meeting of the Association for Computational Linguistics",
    month = jul,
    year = "2020",
    address = "Online",
    publisher = "Association for Computational Linguistics",
    url = "https://aclanthology.org/2020.acl-main.420/",
    doi = "10.18653/v1/2020.acl-main.420",
    pages = "4609--4622"
}

@inproceedings{see-etal-2017-get,
    title = "Get To The Point: Summarization with Pointer-Generator Networks",
    author = "See, Abigail  and
      Liu, Peter J.  and
      Manning, Christopher D.",
    editor = "Barzilay, Regina  and
      Kan, Min-Yen",
    booktitle = "Proceedings of the 55th Annual Meeting of the Association for Computational Linguistics (Volume 1: Long Papers)",
    month = jul,
    year = "2017",
    address = "Vancouver, Canada",
    publisher = "Association for Computational Linguistics",
    url = "https://aclanthology.org/P17-1099/",
    doi = "10.18653/v1/P17-1099",
    pages = "1073--1083"
}

@inproceedings{goldman-etal-2022-un,
    title = "(Un)solving Morphological Inflection: Lemma Overlap Artificially Inflates Models' Performance",
    author = "Goldman, Omer  and
      Guriel, David  and
      Tsarfaty, Reut",
    editor = "Muresan, Smaranda  and
      Nakov, Preslav  and
      Villavicencio, Aline",
    booktitle = "Proceedings of the 60th Annual Meeting of the Association for Computational Linguistics (Volume 2: Short Papers)",
    month = may,
    year = "2022",
    address = "Dublin, Ireland",
    publisher = "Association for Computational Linguistics",
    url = "https://aclanthology.org/2022.acl-short.96/",
    doi = "10.18653/v1/2022.acl-short.96",
    pages = "864--870"
}

@inproceedings{zhang-2023-pronunciation,
    title = "Pronunciation Ambiguities in {J}apanese Kanji",
    author = "Zhang, Wen",
    editor = "Gorman, Kyle  and
      Sproat, Richard  and
      Roark, Brian",
    booktitle = "Proceedings of the Workshop on Computation and Written Language (CAWL 2023)",
    month = jul,
    year = "2023",
    address = "Toronto, Canada",
    publisher = "Association for Computational Linguistics",
    url = "https://aclanthology.org/2023.cawl-1.7/",
    doi = "10.18653/v1/2023.cawl-1.7",
    pages = "50--60"
}

@article{yang-wilcox-arnett-2026,
    title = "Apples to Apples? {T}owards Comparable Crosslingual
             Language Model Evaluation",
    author = "Yang, Xiulin and
              Wilcox, Ethan Gotlieb and
              Arnett, Catherine",
    journal = "arXiv preprint arXiv:2608.25089",
    year = "2026",
    url = "https://arxiv.org/abs/2608.25089"
}

@article{zhang-2026-irregularity,
    title = "When Irregularity Helps: {A} Subclass 
             Analysis of Inductive Bias in Neural 
             Morphology",
    author = "Zhang, Wen",
    journal = "arXiv preprint arXiv:2605.20558",
    year = "2026",
    url = "https://arxiv.org/abs/2605.20558"
}

@article{zhang-2026-moras,
    title = "Mind Your Moras: Orthography-Aware Error 
             Analysis of Neural {J}apanese Morphological 
             Generation",
    author = "Zhang, Wen",
    journal = "arXiv preprint arXiv:2605.20043",
    year = "2026",
    url = "https://arxiv.org/abs/2605.20043"
}

\clearpage
\appendix
\section{Type 4-2 Verb Inventory}
\label{app:type42}

\noindent\parbox{\textwidth}{The following table lists all 37 Type~4-2 verbs in our dataset with their lemma and past-tense forms in \textit{hiragana}.}

\vspace{0.5em}
\begin{center}
\footnotesize
\begin{tabular*}{\textwidth}{@{\extracolsep{\fill}}llll}
\toprule
\multicolumn{2}{c}{\textbf{Verbs 1--19}} & 
\multicolumn{2}{c}{\textbf{Verbs 20--37}} \\
\cmidrule(lr){1-2}\cmidrule(lr){3-4}
\textbf{Lemma} & \textbf{Past Tense} & 
\textbf{Lemma} & \textbf{Past Tense} \\
\midrule
\jp{はねかえる} & \jp{はねかえった} &
\jp{おもいふける} & \jp{おもいふけった} \\
\jp{にげかえる} & \jp{にげかえった} &
\jp{ねがえる} & \jp{ねがえった} \\
\jp{ひっくりかえる} & \jp{ひっくりかえった} &
\jp{ねそべる} & \jp{ねそべった} \\
\jp{あきれかえる} & \jp{あきれかえった} &
\jp{あざける} & \jp{あざけった} \\
\jp{ける} & \jp{けった} &
\jp{むせかえる} & \jp{むせかえった} \\
\jp{よみふける} & \jp{よみふけった} &
\jp{しゃべる} & \jp{しゃべった} \\
\jp{みかえる} & \jp{みかえった} &
\jp{そりかえる} & \jp{そりかえった} \\
\jp{うらがえる} & \jp{うらがえった} &
\jp{あまがける} & \jp{あまがけった} \\
\jp{よみがえる} & \jp{よみがえった} &
\jp{さえかえる} & \jp{さえかえった} \\
\jp{しげる} & \jp{しげった} &
\jp{はべる} & \jp{はべった} \\
\jp{ひるがえる} & \jp{ひるがえった} &
\jp{かげる} & \jp{かげった} \\
\jp{おさまりかえる} & \jp{おさまりかえった} &
\jp{わかがえる} & \jp{わかがえった} \\
\jp{せる} & \jp{せった} &
\jp{すりへる} & \jp{すりへった} \\
\jp{たちかえる} & \jp{たちかえった} &
\jp{おいしげる} & \jp{おいしげった} \\
\jp{いきかえる} & \jp{いきかえった} &
\jp{にえかえる} & \jp{にえかえった} \\
\jp{あせる} & \jp{あせった} &
\jp{ほてる} & \jp{ほてった} \\
\jp{わきかえる} & \jp{わきかえった} &
\jp{まがりくねる} & \jp{まがりくねった} \\
\jp{せせる} & \jp{せせった} &
\jp{つねる} & \jp{つねった} \\
\jp{ひねる} & \jp{ひねった} &
 &  \\
\bottomrule
\end{tabular*}
\end{center}

\vspace{0.5em}
\noindent\parbox{\textwidth}{\small Table A1: All 37 Type~4-2 verbs in our dataset, with their lemma and past-tense forms in \textit{hiragana}.}

\end{document}